\documentclass{article}
\usepackage[preprint]{neurips_2025}

\usepackage[utf8]{inputenc}
\DeclareUnicodeCharacter{0301}{\'{}}
\DeclareUnicodeCharacter{030C}{\v{}}
\usepackage[utf8]{inputenc} 
\usepackage[T1]{fontenc}    
\usepackage{hyperref}       
\usepackage{url}            
\usepackage{booktabs}       
\usepackage{amsfonts}       
\usepackage{nicefrac}       
\usepackage{microtype}      
\usepackage{xcolor}         
\usepackage{graphicx}
\usepackage{amsmath}
\usepackage{multirow}       
\usepackage{subcaption,tikz}
\usepackage{wrapfig}
\usepackage{algorithm,algorithmicx,algpseudocode}
\title{Object Navigation with Structure-Semantic 
 Reasoning-Based Multi-level Map and Multimodal Decision-Making LLM
}

\author{%
  Chongshang Yan\\
  School of Automation\\
  Beijing Institute of Technology\\
  \texttt{3120240840@bit.edu.cn} \\
  \And
  Jiaxuan He\\
    School of Automation\\
  Beijing Institute of Technology\\
    \texttt{1120213032@bit.edu.cn} \\
  \And
  Delun Li\\
    School of Automation\\
  Beijing Institute of Technology\\
    \texttt{3220231166@bit.edu.cn} \\
      \And
  Yi Yang\\
    School of Automation\\
  Beijing Institute of Technology\\
    \texttt{yang\_yi@bit.edu.cn} \\
      \And
  Wenjie Song\\
    School of Automation\\
  Beijing Institute of Technology\\
    \texttt{songwj@bit.edu.cn} \\
}

\begin{document}

\maketitle

\begin{abstract}
The zero-shot object navigation (ZSON) in unknown open-ended environments coupled with semantically novel target often suffers from the significant decline in performance due to the neglect of high-dimensional implicit scene information and the long-range target searching task.
To address this, we proposed an active object navigation framework with Environmental Attributes Map (EAM) and MLLM Hierarchical Reasoning module (MHR) to improve its success rate and efficiency. EAM is constructed by reasoning observed environments with SBERT and predicting unobserved ones with Diffusion, utilizing human space regularities that underlie object-room correlations and area adjacencies. MHR is inspired by EAM to perform frontier exploration decision-making, avoiding the circuitous trajectories in long-range scenarios to improve path efficiency.
Experimental results demonstrate that the EAM module achieves 64.5\% scene mapping accuracy on MP3D dataset, while the navigation task attains SPLs of 28.4\% and 26.3\% on HM3D and MP3D benchmarks respectively - representing absolute improvements of 21.4\% and 46.0\% over baseline methods.
\end{abstract}
\section{Introduction}
\label{sec:intro}
The ZSON task requires agents to locate semantically novel targets in unknown open-ended environments, presenting two key challenges. First, human-centric spaces exhibit multi-level coupled high-dimensional implicit regularities: object distributions adhere to latent functional zoning patterns (e.g., strong kitchen-utensil correlations), while functional zones form topological network structures (e.g., bedroom-bathroom adjacency). When navigating across multiple functional zones (e.g., locating a kitchen target from a bedroom starting point), agents face implicit interference from complex spatial dependencies. Second, long-range exploration encounters the contradiction between temporal information accumulation and real-time decision-making: relying solely on current perceptual snippets causes spatial cognition degradation due to historical context loss, while retaining full temporal information introduces redundant noise, leading to decision-dimensionality catastrophes.

\begin{figure}
    \centering
    \includegraphics[width=1\linewidth]{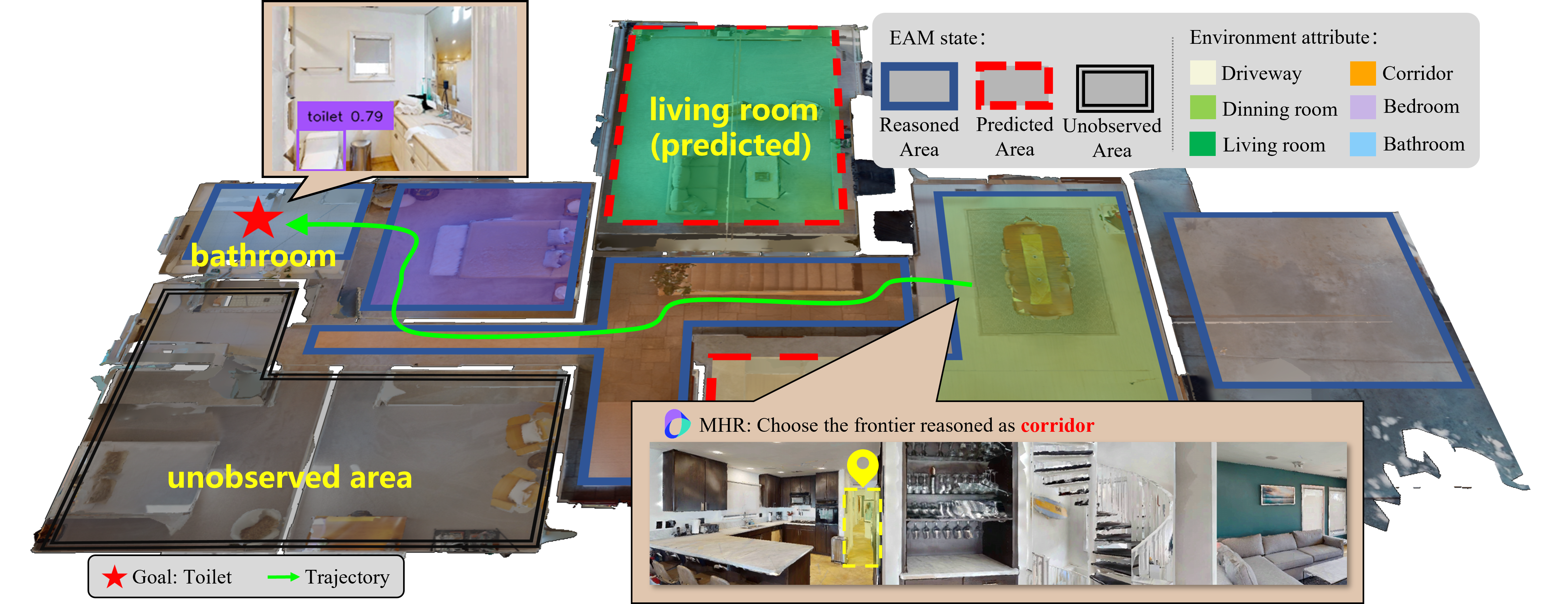}
    \caption{A schematic diagram of navigation using a real-time constructed EAM in an unknown human space, with the navigation target being bathtub. The agent inferences and predicts the surrounding environment properties in real time and navigation process leverages MHR for hierarchical decision-making, starting from the dining room, passing through the corridor, bedroom and finally to the bathroom, to bathtub
}
    \label{fig:main_eg}
\end{figure}

The challenge's complexity stems from three inherent factors: (1) Environmental attributes manifest as implicit high-dimensional signals requiring joint interpretation of cross-modal cues (spatial layouts and object semantics). Simplified approaches like semantic segmentation or rule-based room classification fail to capture these nonlinear correlations. (2) Effective utilization of human-centric spaces demands predictive imagination beyond traditional mapping paradigms to reconstruct environmental properties outside immediate perception, where inaccurate predictions risk misleading exploration decisions. (3) Long-horizon multi-step decision-making necessitates large models' capabilities in persistent memory and multimodal comprehension.

Previous work has demonstrated the efficacy of leveraging human spatial regularities for navigation. Methods such as \cite{xu2024hierarchicalspatialproximityreasoning} (exploiting zonal adjacency relationships), \cite{floorplanguided} aligning prior floor plans with real-time perception), and \cite{zhou2023escexplorationsoftcommonsense}(room-object reasoning via pretrained commonsense language models) have improved path efficiency. Building on these, we posit that explicitly constructing a environment attribute map(EAM) through real-time perception fused with commonsense prediction can unify multi-frame environmental representations. Simultaneously, employing multimodal large language models (MLLMs) to make decisions based on this abstracted map addresses spatial cognition degradation and information redundancy, offering a solution to circuitous paths in long-range navigation.

Our core innovation lies in integrating a Reasoning-Predicting fused map with MHR as shown in Figure \ref{fig:main_eg}, achieving three breakthroughs: 1) Probabilistic spatial completion via diffusion models simulates human-like mental mapping of unobserved regions; 2) SBERT-based real-time reasoning encodes functional spatial correlations into an environmental attribute map (EAM), achieving 64.5\% scene understanding accuracy on MP3D; 3) A multi-modal hierarchical reasoning architecture enables single-instruction multi-round decision-making. Experimental results demonstrate that the navigation task attains SPLs of 28.4\% and 26.3\% on HM3D\cite{hm3d} and MP3D benchmarks respectively. However, performance degrades significantly in non-standard spaces lacking explicit functional zoning (e.g., studio apartments).
\section{Related Work}

As outlined in Section \ref{sec:intro}, ZSON faces two constraints: precise interpretation of environmental semantics and sustained efficiency in long-range exploration. Therefore, recent research has focused on two critical directions accordingly: — (1) environmental understanding, and (2)decision architectures for planning.

\subsection{Environmental Understanding}

Environmental understanding constitutes the cornerstone of zero-shot object navigation. Recent advances in semantic-enhanced representation learning have progressively revealed the critical role of hierarchical scene reasoning capabilities.
Early approaches focused on encoding explicit semantically correlations through structured priors—Li et al. \cite{floorplanguided} demonstrated navigation improvements using manually annotated floorplans, while Pla3D \cite{pla3d} automated cross-modal associations between 3D geometries and textual summaries.  \cite{zhou2023escexplorationsoftcommonsense} leveraged vision-language models for direct attribute inference. \cite{xu2024hierarchicalspatialproximityreasoning} established object-scene co-occurrence strategies, evolved into statistical reasoning paradigms. These semantic-driven methods excel at interpreting observed environments but remain constrained by perceptual boundaries.

The emergence of generative models \cite{LatentDiffusion, nichol2022imagegeneration,Blended} introduced predictive capabilities beyond immediate perception. Diffusion-based approaches \cite{housediff,MSD} revolutionized spatial imagination through probabilistic denoising. RePaint \cite{repaint} achieved coherent image completion through iterative refinement, while architectural applications \cite{latentpaint} demonstrated plausible room layout generation from partial observations. However, such purely generative paradigms face functional-semantic misalignment: although effective at reconstructing geometric structures (as evidenced in \cite{fpdb}), they exhibit inconsistencies in maintaining object-function correspondence, such as misplaced objects across functional zones. 

\subsection{Zero-shot Object Navigation}
Early Foundations in Zero-Shot Object Navigation (ZSON) research builds upon vision-language navigation (VLN) frameworks that explore joint visual-linguistic representations. For instance, \cite{huang2023visuallanguagemapsrobot,xu2024hierarchicalspatialproximityreasoning,yokoyama2023vlfmvisionlanguagefrontiermaps} validated the effectiveness of vision-language alignment in open-vocabulary scene understanding, establishing foundational insights for ZSON. Advances in LLM/MLLM-Driven ZSON Recent progress leverages large language models (LLMs) and multimodal large language models (MLLMs) for their commonsense reasoning and multimodal understanding. Two primary directions have emerged:

Sequential Memory and Multi-Step Decision Optimization. These works focus on LLMs' long-horizon reasoning capabilities by decomposing continuous-space navigation into sequential decision-making across discrete nodes. For example, \cite{voronav} proposed a Voronoi-based hierarchical planner that dynamically evaluates semantic contexts at branching nodes to generate interpretable exploration policies. \cite{zhou2023escexplorationsoftcommonsense} introduced soft constraints to convert commonsense rules (e.g., "kitchens often adjoin dining areas") into probabilistic exploration objectives, optimizing path selection efficiency. Modular frameworks like those in \cite{tan2025mobilerobotnavigationusing,xu2024drivegpt4interpretableendtoendautonomous,humanspacerelationship} integrate LLMs as reusable components to exploit their memory retention across multi-turn interactions.

Multimodal Spatial Representation and Top-View Reasoning This direction addresses spatial information loss caused by vision-to-language conversion in traditional methods. \cite{zhang2024tagmaptextbasedmap} proposed Tag Maps, which generate explicit semantic labels via multi-label classification models for efficient scene context encoding with minimal memory overhead. \cite{topv} advanced this paradigm with the TopV-Nav framework, constructing semantically enriched top-view maps. Its Adaptive Visual Prompt Generation mechanism enables MLLMs to parse room layouts and object spatial relationships directly from top-view perspectives, circumventing information loss from linguistic abstraction.

\section{Our Method}
\subsection{Problem Definition}
\begin{figure}
    \centering
    \includegraphics[width=1\linewidth]{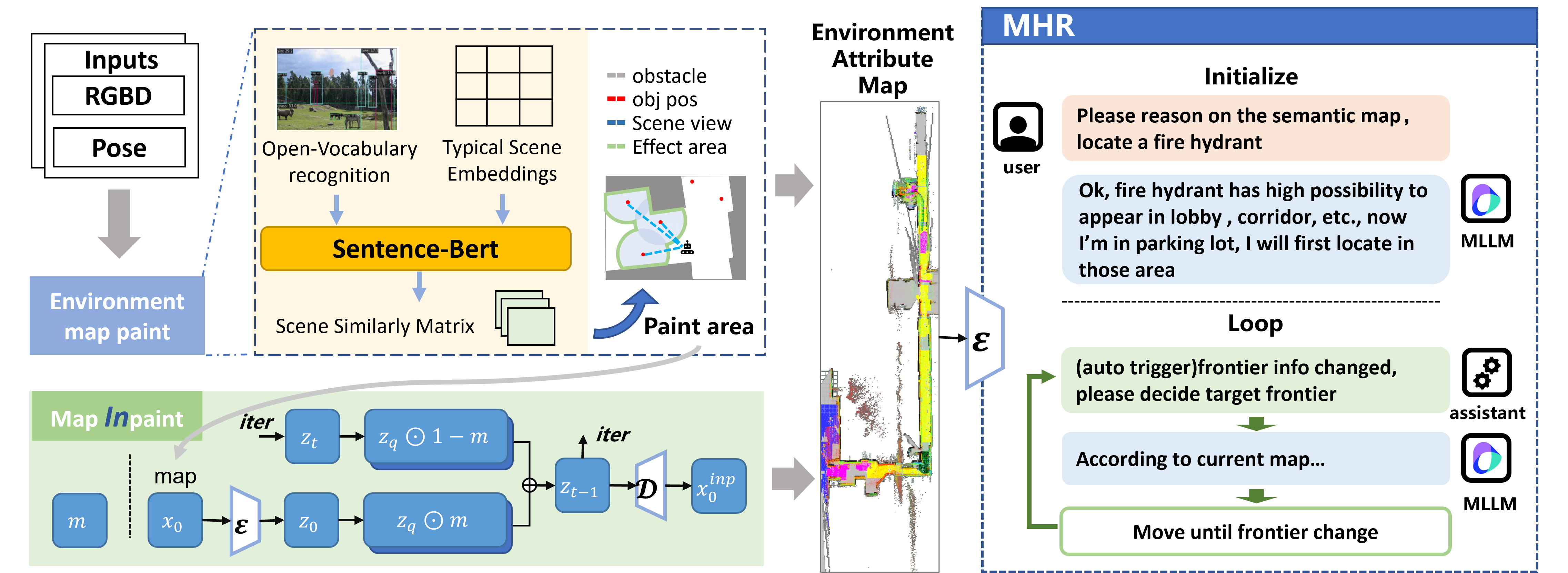}
    \caption{The pipeline of the proposed navigation framework. Observed RGB-D inputs are processed through SBERT-based semantic grounding and diffusion-model-driven spatial imagination. These complementary representations are probabilistically fused to construct EAMs. With the EAMs, the MHR module performs hierarchical spatial-semantic reasoning to formulate navigation policies}
    \label{fig:mainframe}
\end{figure}
In long-range navigation tasks, an MLLM-equipped agent leverages an EAM to make sequential decisions, guiding the robot toward regions with higher spatial likelihood of containing the target object while ultimately achieving goal localization. The task is formally defined as follows:

At initialization, an agent is randomly deployed in an unexplored environment and receives a semantic navigation instruction specifying the target object (e.g., "locate a fire extinguisher"). At time step $t$, the agent receives depth image observation $D_t$ and RGB image observation $C_t$, then makes exploration decisions. The cost function is defined as:
\begin{equation}
    C_{eval} =\frac{\sum^{T}_{t=1}P_a(p_t,g)}{P_s(p_0,g)}
\end{equation}

\subsection{Diffusion-Based Floor-Plan Inpainting}

The diffusion-based floor plan prediction framework comprises three core components: (1) latent space denoising via diffusion processes, (2) diffusion-driven image inpainting, and (3) low-rank adaptive fine-tuning\cite{lora} (LoRA) for domain-specific generation while preserving pre-trained knowledge.

The Stable Diffusion operates in a compressed latent space, where the diffusion process gradually corrupts input data by adding Gaussian noise over iterative steps. For an input image \(x_0\), the noised version at step \(t\) is formulated as:
\begin{equation}
    x_t = \sqrt{\alpha_t} x_{t-1} + \sqrt{1-\alpha_t} \epsilon_t, \quad \epsilon_t \sim \mathcal{N}(0, I)
\end{equation}

where \(\alpha_t\) denotes the cumulative noise coefficient, and \(x_T\) approximates a standard Gaussian distribution after \(T\) steps.

We employ a Variational Autoencoder\cite{VAE} (VAE) to map input floor plans into a compressed latent space, reducing computational complexity while preserving structural details. During conditional generation, text prompts are encoded by a CLIP\cite{clip} text encoder into embeddings, which interact with the U-Net architecture via cross-attention mechanisms. The U-Net, central to Stable Diffusion, predicts noise residuals iteratively:

\begin{equation}
p(x_{i-1} | x_i, \epsilon) = \mathcal{N}\left(x_{i-1}; \mu_\theta(x_i, c, t), \Sigma_\theta(x_i, c, t)\right)
\end{equation}

where \(\mu_\theta\) denotes the noise predictor conditioned on the noised image \(x_i\), text embedding \(c\), and timestep \(t\).

For partial observations, let \(x\) denote the complete image, \(m\) the unknown region mask, and \(1-m\) the observed area. We adopt an iterative inpainting strategy that replaces masked regions while preserving unmasked content. At each reverse step:

\begin{equation}
\gamma \sim \mathcal{N}\left(\sqrt{\alpha_t} x_{t-1}^{known}, (1-\alpha_t)I\right)
\end{equation}

\begin{equation}
\mu(x_{t-1}^{unknown}, t), \quad \Sigma(x_{t-1}^{unknown}, t)
\end{equation}

\begin{equation}
x_{t-1} = m \odot x_{t-1}^{unknown} + (1-m) \odot x_{t-1}^{known}
\end{equation}

Here, \(x_{t-1}^{known}\) is sampled from the observed region, while \(x_{t-1}^{unknown}\) is predicted by the U-Net.

While pre-trained VAEs reconstruct natural images effectively, they produce artifacts when applied to architectural floor plans. We partially fine-tune the first ResNet block in the encoder and the last ResNet block in the decoder—components directly responsible for low-level feature extraction and high-level reconstruction—to eliminate structural inconsistencies.

\subsection{Environmental Attributes Mapping}
\begin{figure}
    \centering
    \includegraphics[width=1\linewidth]{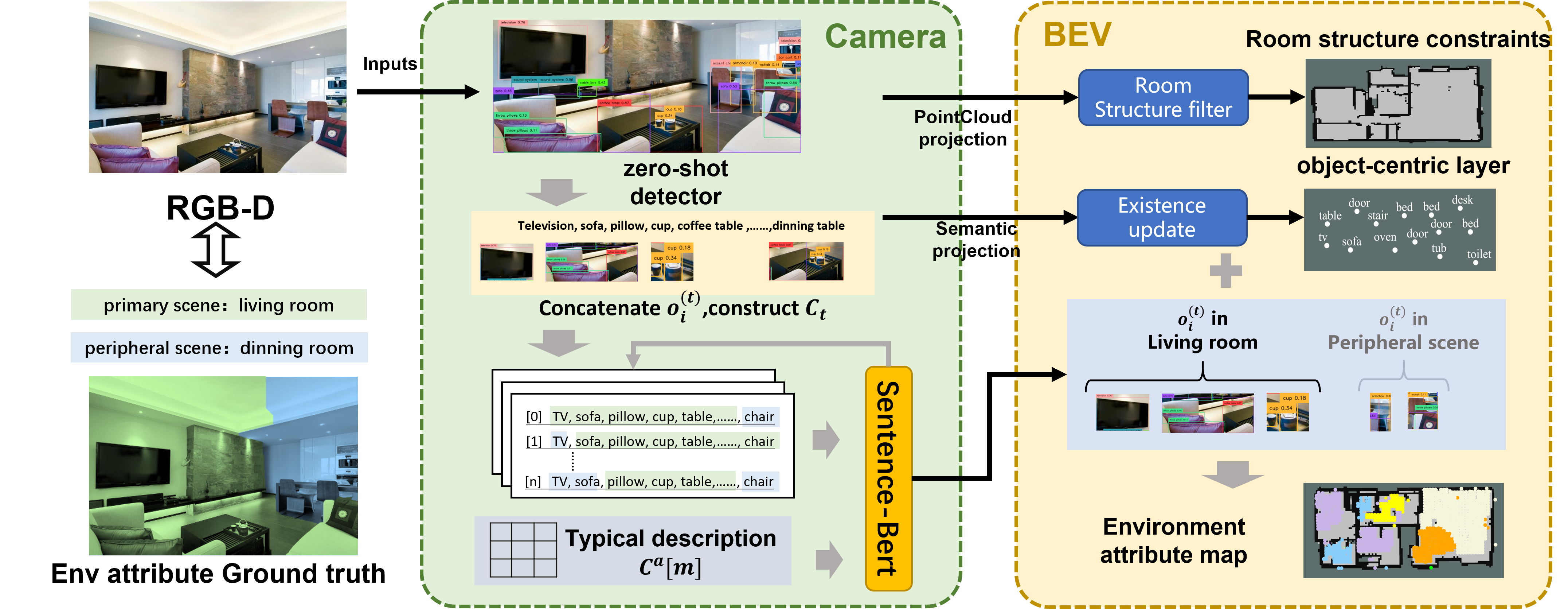}
    \caption{Environmental Attributes Mapping framework. RGB-D observations are projected into 3D point clouds. SBERT fine-tuned on HM3D triplet data grounds semantic attributes from zero-shot object detections, propagating contextual properties under geometric constraints, enabling hierarchical semantic-geometric fusion. Cross-layer alignment synthesizes object distributions and structural embeddings into a unified 2D EAM for spatially informed navigation.}
    \label{fig:eam_frame}
\end{figure}
Our approach integrates multi-modal sensory inputs into a structured 2D grid-based map as shown in Figure\ref{fig:eam_frame}. Specifically, the map comprises three distinct layers: (1) an object-centric layer that captures detected object categories and their spatial distributions, (2) an environmental attribute layer that infers scene properties from object spatial patterns and semantic correlations, and (3) an exploration layer that tracks observed and unexplored regions while dynamically generating boundary attributes based on environmental layer insights and general spatial patterns.

For each step, acquire RGB-D observation $\mathcal{F}_t = (\mathcal{I}_t, \mathcal{D}_t) \in \mathbb{R}^{H\times W\times 4}$ where $\mathcal{I}_t$ is RGB image and $\mathcal{D}_t$ depth map and agent pose $P_t = (R_t, t_t) \in \text{SE(3)}$ consists of rotation $R_t \in \text{SO(3)}$ and translation $t_t \in \mathbb{R}^3$. To build these layers, $\mathcal{D}_t$ and $P_t$ are first utilized to generate 3D point clouds. Points near ground level are projected onto the exploration layer to delineate traversable regions, while elevated points are mapped to the environmental attribute layer as room structure constraints. Then apply zero-shot detector on $I_t$ to construct the scene description $C_t$:

\begin{equation}
C_t = \left[\; o^{(t)}_1 \oplus o^{(t)}_2 \oplus \cdots \oplus o^{(t)}_{m_t} \;\right], \quad o^{(t)}_i \sim \text{Det}_{\text{zs}}(\mathcal{I}_t)
\end{equation}

where $\oplus$ represents semantic object concatenation, and $\text{Det}_{\text{zs}}$ denotes the zero-shot detector. The description corpus $\mathcal{C} = \{C_i\}_{i=1}^M$ aggregates observations across temporal windows.

To enhance the classification of environmental attributes, we fine-tuned the Sentence-BERT\cite{SBERT} (SBERT) model on HM-3D for inferring scene properties.
Given a 3D scene $S \in \mathbb{S}_{\text{HM3D}}$, we sample $K$ camera poses $\{P_k\}_{k=1}^K$ where each.  Process $\mathcal{I}_k$ to construct scene description $C_k$ as illustrated above. We optimize SBERT parameters $\theta$ through domain adaptation:

Define the triplet components as:
\begin{itemize}
\item Anchor ($C^a$): Canonical scene descriptions from HM-3D/MP-3D
\item Positive ($C^+$): Descriptions with base SBERT accuracy $\alpha(C^+) \geq 0.8$
\item Negative ($C^-$): Descriptions with $\alpha(C^-) \leq 0.2$
\end{itemize}

Each object $o_i$ in HM-3D scenes is pre-annotated with environmental attribute $e_i^* \in \{e_{\text{kitchen}}, e_{\text{office}}, ...\}$. This provides per-instance supervision signals for fine-tuning. The scene description accuracy is computed as:
\begin{equation}
\alpha(C) = \frac{1}{|C|} \sum_{o \in C} \delta(f_{\theta_0}(o), e^*_o),
\end{equation}
where $\delta(a,b) = 
\begin{cases}
1 & \text{if } a = b \\
0 & \text{otherwise}
\end{cases}
$ is the binary agreement function.

We formulate a scene parsing framework where a single RGB observation contains a primary scene and peripheral scenes. Subsequently, random $o_i$ are aggregated to construct scene descriptions $C$. A greedy algorithm optimizes these descriptions by iteratively adjusting object combinations to maximize similarity with canonical scene templates, effectively filtering out objects incongruent with the primary scene context.

We subsequently develop environmental attributes by propagating contextual properties from the object-centric origin under room layout constraints, progressively expanding to encompass the surrounding spatial configuration. Formally, let the grid-based map $\mathcal{G} \in \mathbb{R}^{H \times W \times 3}$ represent fused environmental attributes, where each grid cell $\mathcal{G}_{(i,j)} = [o_{(i,j)}, e_{(i,j)}]$ contains:
\begin{itemize}
    \item \textbf{Object distribution} $o_{(i,j)} \in [0,1]^K$: Probability vector of $K$ object categories
    \item \textbf{Environmental attribute} $e_{(i,j)} \in \mathbb{R}^d$: $d$-dimensional embedding inferred from SBERT
\end{itemize}

The propagation process implements iterative neighborhood aggregation:
\begin{equation}
    e_{(i,j)}^{t+1} = e_{(i,j)}^t + \sum_{(k,l)\in \mathcal{N}(i,j)} w_{kl}^t e_{(k,l)}^t
\end{equation}
where $\mathcal{N}(i,j)$ denotes 8-connected neighbors and $w_{kl}^t$ weights derived from room layout geometry.
\subsection{MLLM Hierarchical Reasoning module}
\label{ss:MLLM}

\begin{wrapfigure}[19]{r}[3pt]{6cm}
    \centering
    \includegraphics[width=1\linewidth]{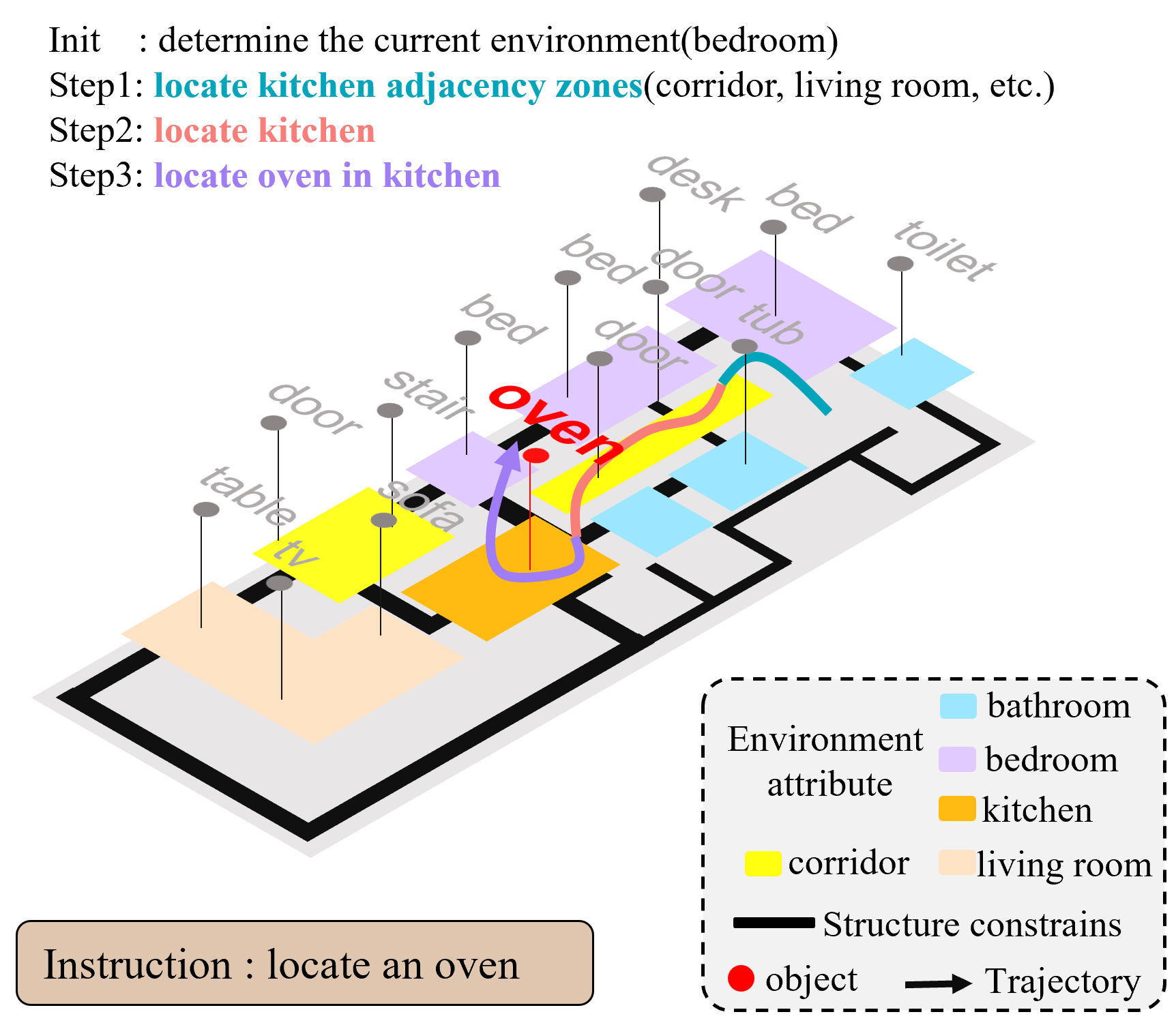}
    \caption{Multi-Layer Environmental Map Navigation Example}
    \label{fig:nav_eg}
\end{wrapfigure}

To enhance embodied agents' target-driven navigation capabilities in unseen environments, we develop a multimodal hierarchical reasoning architecture (MHR) using Doubao-vision-pro-32k\cite{doubao}  that emulates human cognitive reasoning as illustrated in Figure \ref{fig:nav_eg}. 
When navigating toward an oven, the MLLM hierarchically guides the agent by: 1) prioritizing regions with high spatial adjacency likelihood to kitchens based on scene attribute priors, 2) exploring identified kitchen zones, and 3) conducting local object search within the target area. This hierarchical strategy effectively bridges semantic understanding and geometric exploration.

Our methodology synergizes LLM-driven semantic reasoning with boundary-aware exploration. We first extract structural edges. These edges are then enriched with scene attributes through cross-modal fusion, generating an attribute-augmented exploration map. The edge embeddings provide spatial referential cues for the MLLM's decision-making module.

To mitigate computational latency from sequential LLM reasoning while maintaining exploration efficiency, we implement a hybrid planning approach: 1) A Traveling Salesman Problem (TSP) solver optimizes the visitation order of high-value edges based on their exploration scores and spatial proximities. 2) Local waypoints are dynamically sampled along the globally optimized path for reactive navigation. This two-tiered planning strategy ensures real-time responsiveness without compromising the systematic exploration guided by semantic understanding.

\section{Performance Experiments}
\label{sec:exp}
\subsection{Experimental Setup}
\textbf{Test Environments.}
We evaluate our framework on HM3D datasets under the Habitat-Sim simulator\cite{habitat19iccv,szot2021habitat,puig2023habitat3}, focusing on long-range navigation tasks (>15m initial-target distances). 

\textbf{Evaluation Metrics.}
We adopt Success Rate(SR) and Success weighted by Path Length (SPL) as the evaluation metrics.Specifically, we introduce \textit{Scene Understanding Consistency} (SUC) and \textit{effective prediction percentage} (EPP) as two evaluation metrics for scene attribute-related modules. The metrics are defined as follows:

\begin{itemize}
  \item \textit{Success Rate (SR)}: Measures the proportion of successfully completed navigation tasks across all test scenarios
  \item \textit{Success weighted by Path Length (SPL)}: Evaluates the agent's locomotion effectiveness in goal-oriented navigation through a standardized ratio comparing the actual traversed path length to the theoretically optimal path length.
  \item \textit{Scene Understanding Consistency (SUC)}: Semantic similarity between predicted and ground-truth regions
  \item \textit{effective prediction percentage (EPP)}: 
  The ratio of the predicted correct area size to the observed area size.
\end{itemize}

\textbf{Baselines.}
Two representative baselines are compared:

\begin{enumerate}
  \item \textbf{Frontier}: agents systematically detect and navigate along frontier edges to optimize spatial coverage efficiency while minimizing redundant traversal.
  \item \textbf{ESC}: Leverages pre-trained vision-language models for open-world scene understanding and LLM-based commonsense reasoning to efficiently locate target objects.
\end{enumerate}

\subsection{Component Validation}

\begin{figure}[thbp]
    \centering
    \captionsetup[subfigure]{labelformat=simple}
    
    \includegraphics[width=0.85\linewidth]{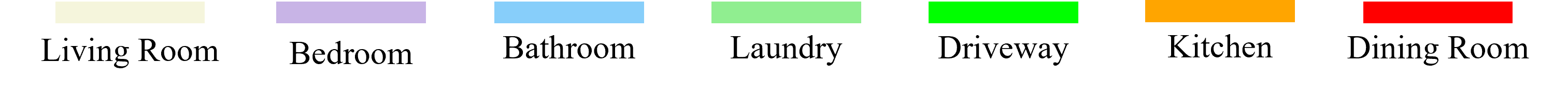}
    
    \begin{minipage}[t]{\textwidth}
        \centering
        \begin{subfigure}[t]{0.49\textwidth}
            \centering
            \begin{tabular}{@{}c@{\hspace{1mm}}c@{\hspace{1mm}}c@{}}
                \includegraphics[width=\linewidth,height=4cm,keepaspectratio]{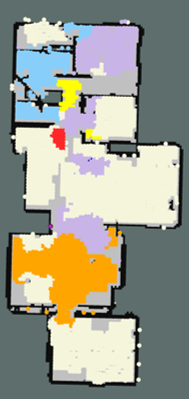} &
                \includegraphics[width=\linewidth,height=4cm,keepaspectratio]{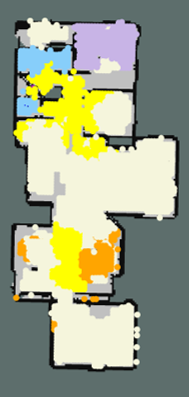} &
                \includegraphics[width=\linewidth,height=4cm,keepaspectratio]{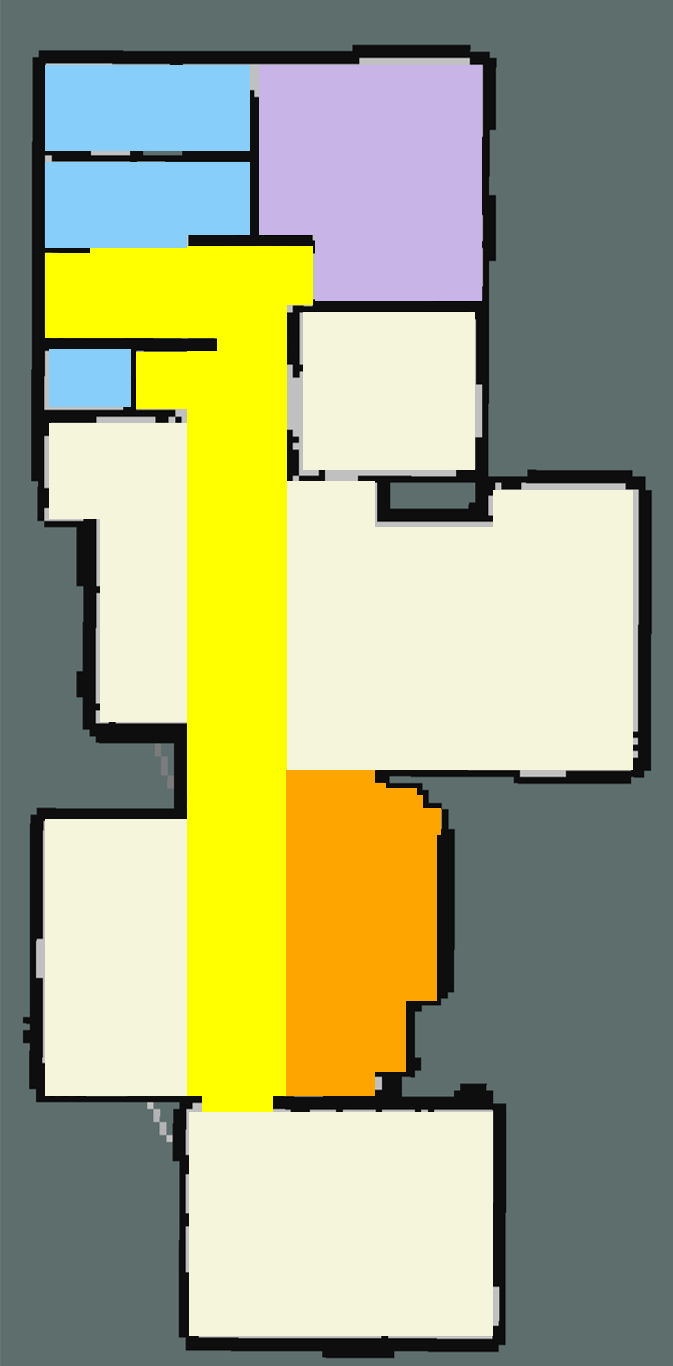} \\
                \footnotesize Base & \footnotesize Tuned & \footnotesize GT
            \end{tabular}
            \label{fig:group_a}
        \end{subfigure}%
        \hfill 
\begin{subfigure}[t]{0.49\textwidth}
            \centering
            \begin{tabular}{@{}c@{\hspace{1mm}}c@{\hspace{1mm}}c@{}}
                \includegraphics[width=\linewidth,height=4cm,keepaspectratio]{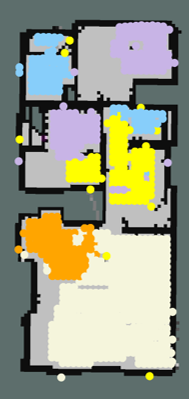} &
                \includegraphics[width=\linewidth,height=4cm,keepaspectratio]{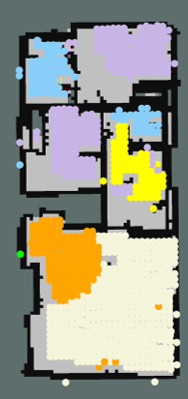} &
                \includegraphics[width=\linewidth,height=4cm,keepaspectratio]{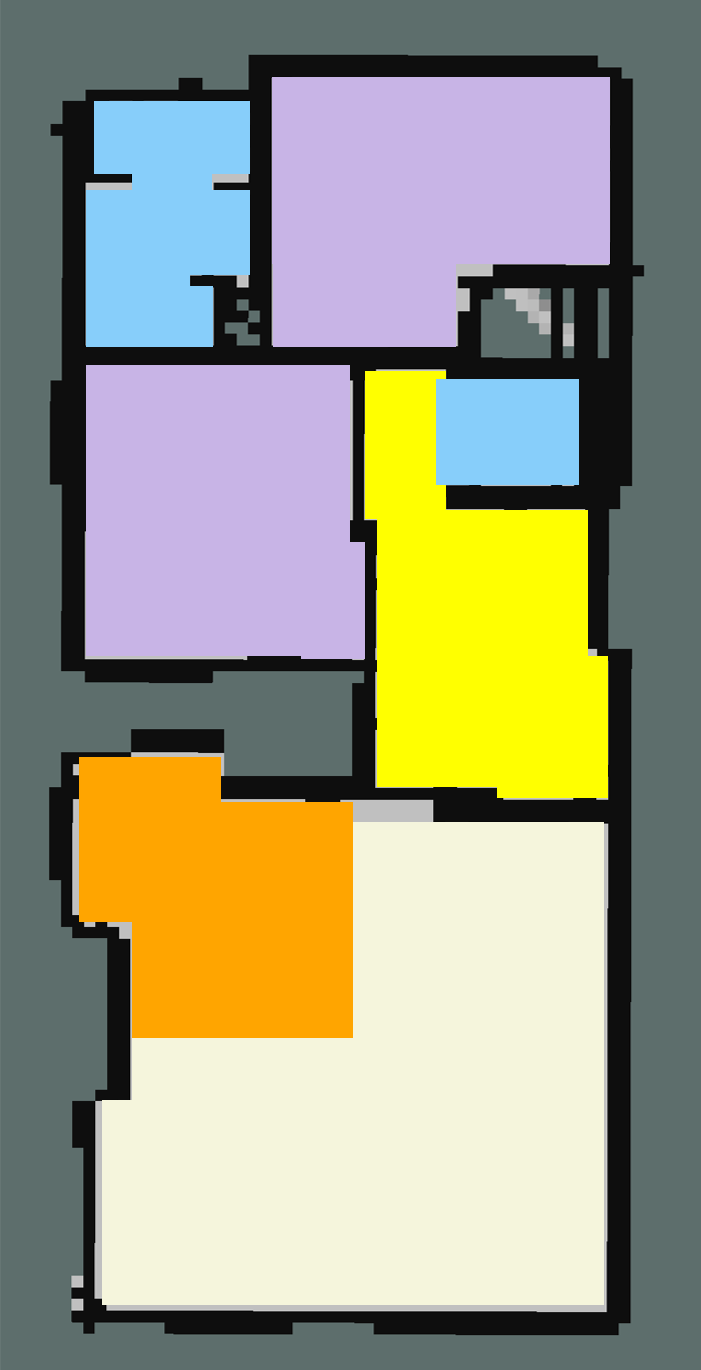} \\
                \footnotesize Base & \footnotesize Tuned & \footnotesize GT
            \end{tabular}
            \label{fig:group_b}
        \end{subfigure}
    \end{minipage}
    


    \caption{Comparative visualization of three representative groups demonstrating performance contrasts among the base model, fine-tuned model, and environmental attribute ground truth (GT). The base model exhibits limited environmental comprehension capability, failing to distinguish primary objects from secondary elements - for instance, misclassifying environments as bedrooms upon detecting multiple pillows.}
    \label{fig:comparison}
\end{figure}
We conducted component evaluations in a closed testing environment without predefined navigation targets. During the exploration phase, we executed EAM module while systematically recording and evaluating the EPP. The agent was allowed to fully explore and observe the environment until comprehensive perceptual coverage was achieved, after which we specifically assessed the SUC of the EAM module. The experimental results are presented in the table \ref{tab:ComponentVal}.

\begin{wraptable}[9]{r}[3pt]{8cm}
    \centering
    \caption{Component Validation}
    \label{tab:ComponentVal}
    \begin{tabular}{@{} l *{4}{c} @{}}
    \toprule
    \multirow{2}{*}{Module} & 
    \multicolumn{2}{c}{HM3D} & \multicolumn{2}{c}{MP3D} \\
    \cmidrule(lr){2-3} \cmidrule(lr){4-5}
    & SUC$\uparrow$ & EPP$\uparrow$ & SUC$\uparrow$ & EPP$\uparrow$ \\
    \midrule
    EAM(base)         & 44.8 & 23.6 & 47.1 & 25.3 \\
    EAM(finetune)       & 51.1 & 35.2 & 64.5 & 36.2 \\
    \bottomrule
    \end{tabular}
\end{wraptable}

The living room area exhibits inherent definitional ambiguity compared to well-structured zones like bedrooms and kitchens, coupled with its substantial spatial extent. These characteristics introduce significant interference in quantitatively assessing fine-tuning performance when evaluating across all residential areas. Systematic exclusion of living room metrics reveals that the fine-tuning procedure demonstrates a 24\% greater enhancement in non-living-room zones compared to evaluations incorporating the complete spatial domain.

\subsection{Comparison With SOTA Methods}
\label{sub:SOTA}
\begin{figure}[htbp]
    \centering
    \includegraphics[width=1.03\textwidth]{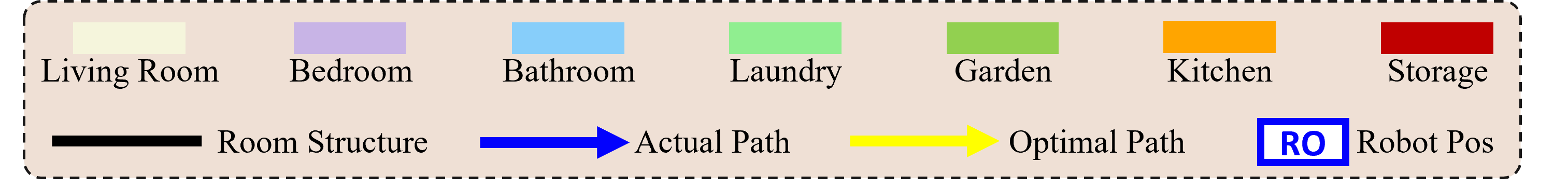}
    \vspace{0mm}
    
    \begin{subfigure}{0.23\textwidth}
        \includegraphics[width=\textwidth]{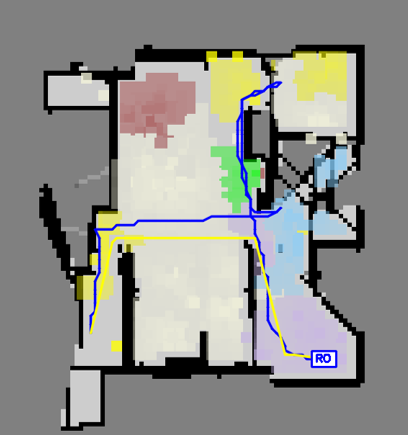}
        \caption{goal:bed}
        \label{fig:nav1}
    \end{subfigure}
    \hfill
\begin{subfigure}{0.23\textwidth}
        \includegraphics[width=\textwidth]{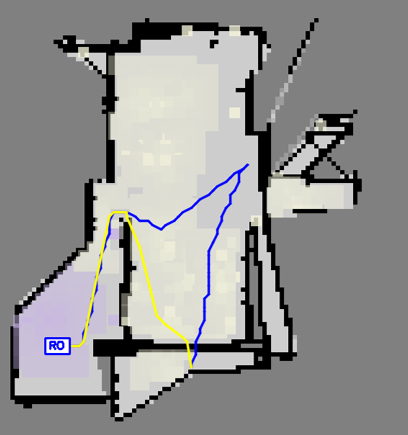}
        \caption{goal:bed}
        \label{fig:nav2}
    \end{subfigure}
    \hfill
\begin{subfigure}{0.23\textwidth}
        \includegraphics[width=\textwidth]{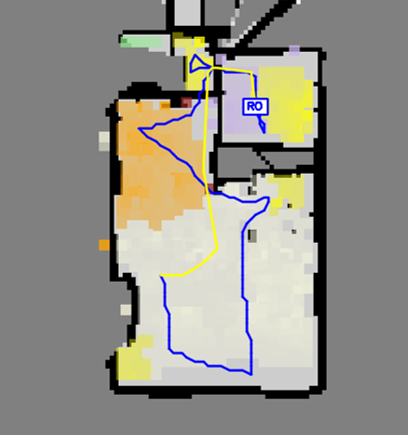}
        \caption{goal:bed}
        \label{fig:nav3}
    \end{subfigure}
    \hfill
\begin{subfigure}{0.23\textwidth}
        \includegraphics[width=\textwidth]{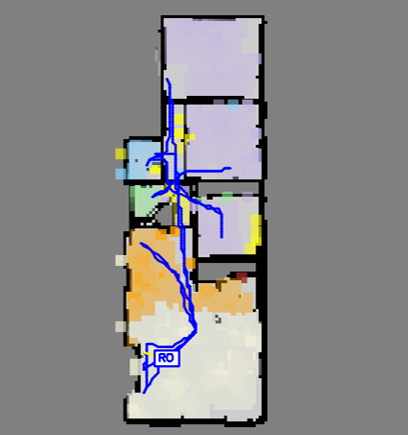}
        \caption{goal:not exist in scene}
        \label{fig:nav4}
    \end{subfigure}
    \caption{Some representative navigation cases which demonstrate the target navigation strategy proposed in our methodology: agents sequentially enter individual spaces for inspection and abandon current space to explore alternative areas when detecting low probability of target existence. This space-switching mechanism enables efficient exploration by dynamically evaluating environmental observations against prior spatial knowledge.}
    \label{fig:navs}
\end{figure}
\begin{table}[h]
    \centering
    \caption{Comparison Result With SOTA Methods}
    \label{tab:Comparison SOTA}
    \begin{tabular}{@{} l *{4}{c} @{}}
    \toprule
    \multirow{2}{*}{Method} & 
    \multicolumn{2}{c}{HM3D} & \multicolumn{2}{c}{MP3D} \\
    \cmidrule(lr){2-3} \cmidrule(lr){4-5}
    & Success$\uparrow$ & SPL$\uparrow$ & Success$\uparrow$ & SPL$\uparrow$ \\
    \midrule
    Frontier  &33.7 & 15.3 & 36.0 & 17.7 \\
    ESC       &39.2 & 22.3 & 28.7 & 14.2 \\
    Ours      & \textbf{43.1} & \textbf{28.4} & \textbf{41.2} & \textbf{26.3} \\ 
    \bottomrule
    \end{tabular}
\end{table}

This paper conducts a systematic evaluation of the proposed zero-shot object navigation framework on the HM3D and MP3D datasets, and compares it with cutting-edge zero-shot methods (such as ESC) and rule-based exploration methods (such as Frontier). The experimental results show that the combination of the Environmental Attribute Map (EAM) and the multimodal hierarchical reasoning module (MHR) proposed in this paper significantly improves the efficiency of long-distance exploration and the cross-scene generalization ability.
 
 On the HM3D dataset, the SR of the method proposed in this paper reaches 43.1\%, which is 9.1\% higher than that of the ESC method based on visual language models (39.2\%) and the Frontier method based on boundary exploration (33.7\%), respectively. SPL reached 28.4\%, increasing by 6.1\% and 13.1\% respectively compared with ESC (21.5\%). This improvement is attributed to the collaborative mechanism of the EAM module through SBERT semantic reasoning and spatial completion of the diffusion model, which enables it to infer the functional attributes of unobserved areas in real time (such as the adjacity relationship between the kitchen and the dining room), thereby reducing redundant exploration paths. For example, in the task of "navigating from the bedroom to the bathroom", EAM guides the MHR module to prioritize the exploration of high-probability areas through the prior spatial topology (the bedroom is usually adjacent to the corridor or changing area), avoiding the detour trajectories caused by the reliance on local perception in traditional methods.

\subsection{Ablation Study}
To validate the efficacy of individual modules, we conducted systematic ablation studies on both HM3D and MP3D datasets. The full model configuration maintained consistency with the implementation details outlined in section \ref{sub:SOTA}, serving as the baseline for comparative analysis against component-disabled variants.

\textbf{Effect of EAM module.}
To validate the effectiveness of our EAM module, we developed two comparative implementations: LLM-EAM and GLIP-EAM. The LLM-EAM variant  employs a large language model to classify detected objects into environmental categories based on linguistic features. In contrast, the GLIP-EAM implementation performs visual scene understanding via the GLIP model.
Experiments show the decline in the efficiency of the comparison method (Table  \ref{tab:EAMAblation}). The response frequency of LM-EAM is relatively lower than that of EAM. There are more uninferred frames, and the judgment error of a single frame may lead to long-term navigation misguidance. Because GLIP-EAM failed to achieve effective scene discrimination in complex scenes with multiple environments in the same field of view, the accuracy of scene recognition decreased.

\begin{table}[htbp]
    \begin{minipage}{\textwidth}
        \begin{minipage}[t]{0.45\textwidth}
                \centering
                \caption{Comparison between different Scene Common Sense Understanding Module}
                \label{tab:EAMAblation}
                \begin{tabular}{@{} l *{3}{c} @{}}
                \toprule
                Module & Success$\uparrow$ & SPL$\uparrow$\\
                \midrule 
                LLM-EAM      & 33.1 & 16.2  \\
                GLIP-EAM     & 36.5 & 22.7  \\
                EAM(Ours)          & 43.1 & 28.4  \\
                \bottomrule
                \end{tabular}
        \end{minipage}
        \hfill
        \begin{minipage}[t]{0.45\textwidth}
                \centering
                \caption{Comparison between different LLM Decision Module}
                \label{tab:MHRAblation}
                \begin{tabular}{@{} l *{3}{c} @{}}
                \toprule
                Module & Success$\uparrow$ & SPL$\uparrow$\\
                \midrule 
                Random        & 32.2 & 14.1  \\
                LLM           & 35.6 & 18.5  \\
                MHR(Ours)          & 43.1 & 28.4  \\
                \bottomrule
                \end{tabular}
        \end{minipage}
    \end{minipage}
\end{table}

\textbf{Effect of HMR module.}
To validate the effectiveness of our MHR, we developed two comparative implementations: Random and LLM. Random prioritizes exploration targets through stochastic environment sampling from the candidate frontier set, while LLM translate explorable frontier into linguistic descriptors serving as exclusive inputs for navigation policy generation.Experiment results are shown in Table \ref{tab:MHRAblation}. Among them, RANDOM generated a success rate similar to that of Frontier. LLM, due to its inability to utilize spatial attributes during decision-making, experienced a decrease in efficiency compared to MHR.
\section{The Conclusions}
\label{sec:conclusion}
This work proposes a zero-shot object navigation framework integrating a Structure-Semantic Reasoning-based EAM with a multimodal hierarchical decision module (MHR). The EAM leverages SBERT-driven semantic grounding and diffusion probabilistic spatial completion, achieving 64.5\% scene understanding consistency on MP3D by modeling complex spatial dependencies. The MHR significantly improves long-range exploration efficiency, yielding absolute SPL improvements of 21.4\% on HM3D and 46.0\% on MP3D benchmarks. Experimental results demonstrate that combining EAM with MHR enhances navigation performance in human-centric environments. However, the framework exhibits limitations in non-structured spaces with ambiguous functional zoning (e.g., studio apartments). Future work will focus on improving adaptability through dynamic spatial-semantic alignment and cross-modal reinforcement learning, while extending the framework via mobile manipulator deployment algorithms to enhance environmental interaction capabilities and broaden real-world applicability.

\newpage
\bibliographystyle{unsrt}
\bibliography{refs}
\newpage
\appendix
\section{Algorithm Specifications}

\label{app:algorithms}

This chapter presents the algorithm specifications of two important modules of our work. Among them, algorithm \ref{alg:eam} detail the EAM module, while algorithm \ref{alg:mhr} presents the MHR module.
\begin{algorithm}
\caption{EAM Generation Algorithm}
\label{alg:eam}
\begin{algorithmic}[1]
\Require
    $F_t = (I_t, D_t)$ \Comment{Current frame RGBD image} \\
    $P_t = (R_t, t_t)$ \Comment{Robot pose} \\
    $\theta$ \Comment{SBERT model parameters} \\
    $G_{prev}$ \Comment{Previous EAM} \\
    $OCC_{prev}$ \Comment{Previous Occupancy Map}
\Ensure
    $G_{curr}$ \Comment{Updated EAM}

\State \textbf{Camera Input}
\State $PC_t \gets \text{ProjectToPointCloud}(D_t, P_t)$ 
\State $O_t \gets \text{ZeroShotDetector}(I_t)$
\State $C_t \gets \bigoplus_{o \in O_t} o$ 
\State $\mathbf{e}_t \gets \text{SBERT}(C_{std};C_t; \theta)$ \Comment{Encode scene attributes}

\State \textbf{EAM Observation Update}
\State $OCC_{curr} \gets \text{update}(OCC_{prev};PC_t)$ \Comment{Probability grid map update}
\For{each grid cell $g \in G_{curr}$}
    \If{$g$ is within observation range}
        \State $g.o \gets \text{ObjectProjection}(PC_t, g)$ 
        \State $g.e \gets \mathbf{e}_t$ 
    \Else
        \State $g.e \gets \text{Unknown}$ 
    \EndIf
\EndFor
\State \textbf{EAM Global Update}
\While{RobotState is Navigation}
    \For{each grid cell $g \in G_{curr}$}
        \If{$g.e \neq \text{Unknown}$}
            \State $\mathcal{N} \gets \text{GetFreeNeighbors}(g;OCC_{curr})$ 
            \State $w \gets \text{CalculateWeights}(\mathcal{N}, g)$ 
            \State $\Delta e \gets \sum_{n \in \mathcal{N}} w_n \cdot n.e$ 
            \State $g.e \gets g.e + \alpha \cdot \Delta e$ \Comment{Update environment attribute}
        \EndIf
    \EndFor
\EndWhile

\State \Return $G_{curr}$
\end{algorithmic}
\end{algorithm}

Note that $g.e$ and $\textbf{e}_t$ are vectors recording various environmental attributes, where the component with the highest value represents the environmental attribute finally reflected in the EAM.

\begin{algorithm}
\caption{MHR Module}
\label{alg:mhr}
    \begin{algorithmic}[1]
        \Require
            $G_t$ \Comment{Current EAM} \\
            $OCC_t$ \Comment{Current Occupancy}\\
            $Q_t$ \Comment{Query object category} \\
            $O_t$ \Comment{Objects detected in the scene} \\
            $P_{robot}$ \Comment{Current robot position} \\
        \Ensure
            $P_{target}$ \Comment{Local Explore Target}

        \If{$Q_t$ not in $O_t$}
            \State \textbf{Input Preprocess}
            \State $FR_{t} \gets Sobel(OCC_t) $
            \State $F_{EAM} \gets \text{PaintEAMPic}(OCC_t;G_t;P_{robot})$ 
            \State $F_{obj} \gets \text{PaintObjPic}(OCC_t;O_t;P_{robot})$
            \State $Prompt_A \gets PromptGeneration(Q_t) $

            \State \textbf{Commensence Level Reasoning}
            \State $E_Q \gets MLLM(Q_t,Prompt_A)$ 
            \State $Prompt_B \gets PromptGeneration(E_Q;FR_{t}) $
            \State $Prompt_C \gets PromptGeneration(E_Q;O_t) $
            
            \State \textbf{Environment Level Reasoning}
            \State $FR_{c}\gets MLLM(F_{EAM};Prompt_B)$ 
            \State $P_{FR} \gets getPosition(FR_c)$
            \State \textbf{Object Level Reasoning}
            \If{$P_{robot}$ in $E_Q$}
                \State $P_{Obj}\gets MLLM(F_{Obj},Prompt_C)$  
            \EndIf

            \State $Path \gets TSPSolver(P_{FR};P_{Obj})$
            \State $P_{target} \gets Sample(Path)$
        \Else
        
        \State $P_{target} \gets P_{Q_t}$
        \EndIf
        \State \Return $P_{target}$
    \end{algorithmic}
\end{algorithm}
\section{Training and Model Configurations}
\subsection{Diffusion Model Training Protocols}

The experiments were conducted based on the RPLAN dataset, which is a large - scale floor plan dataset specifically for East Asian residential buildings. The dataset contains RGB three - channel images with a resolution of 512×512 and encompasses 16 types of building components (such as room areas like living rooms, kitchens, bathrooms; door areas like front doors, interior doors; and boundary areas like walls and exterior backgrounds). Each type of component is annotated with a specific RGB value. The dataset is divided into 50,000 training images, 100 validation images, and 1,000 test images.

We fine-tuned a Stable Diffusion 1.5 model, a classic latent diffusion model (LDM) pre-trained on the LAION-5B dataset (5.8+ billion image-text pairs). Key hyperparameters are listed in Table \ref{tab:diffusion-hyperparameters}:  
\begin{table}[h]
    \centering
        \begin{tabular}{ll|ll}
        \hline
        \textbf{Hyperparameter} & \textbf{Value} & \textbf{Hyperparameter} & \textbf{Value} \\
        \hline
        Image Resolution & 512×512 & Batch Size & 8 \\
        Training & 50,000 & Adam Optimizer $\beta_1$ & 0.9\\
        Validation & 100 & Adam Optimizer $\beta_2$ & 0.999\\
        Test Split & 1000 & Adam Optimizer $\lambda_{wd}$ & 0.01\\
        U-Net Learning Rate & 1 & LoRA Rank & 32\\
        Scene Categories & 16 & & \\
        \hline
        \end{tabular}
    \caption{Diffusion model training hyperparameters}
    \label{tab:diffusion-hyperparameters}  
\end{table}

\subsection{SBERT Fine-Tuning}

Since the training base of SBERT lacks training for object-environment classification tasks, we choose to fine-tune the SBERT base to endow SBERT with general knowledge of environment classification.We employ \texttt{MultipleNegativesRankingLoss} to maximize similarity between anchor-positive pairs while distancing negatives. For batch $B$ containing $n$ triplets:

$$ \mathcal{L} = -\frac{1}{n}\sum_{i=1}^{n}\log\frac{e^{\langle a_i, p_i \rangle}}{\sum_{j=1}^{n}e^{\langle a_i, p_j \rangle}} $$

where $\text{sim}(\cdot)$ denotes cosine similarity. Validation employs a \texttt{TripletEvaluator} measuring:
\begin{itemize}
    \item Anchor-Positive similarity: Maintains $>0.85$ throughout training
    \item Anchor-Negative margin: Achieves $>0.2$ separation threshold
\end{itemize}

\begin{figure}[htbp]
    \centering
    \includegraphics[width=1\linewidth]{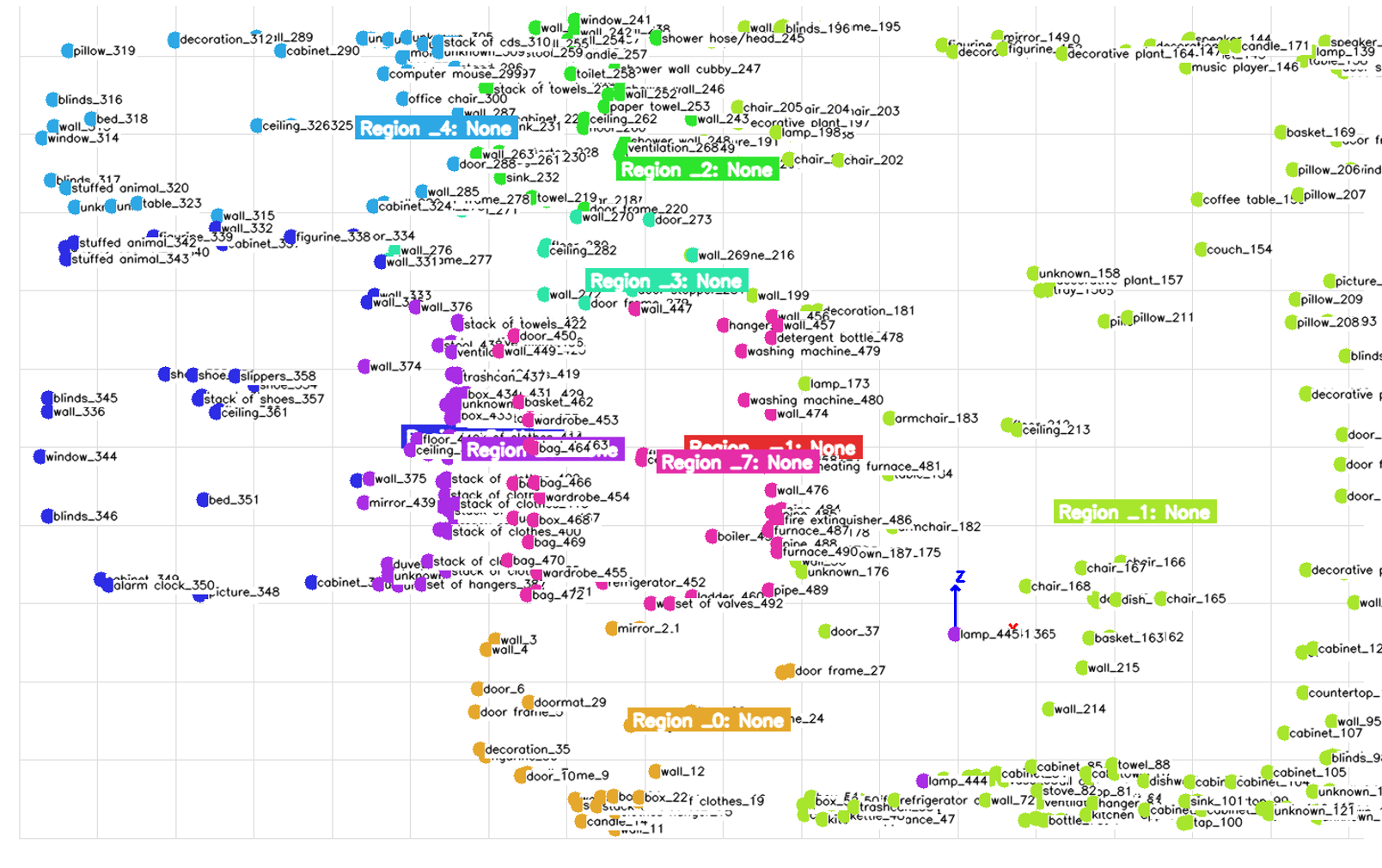}
    \caption{Object-Region relationship ground truth}
    \label{fig:regiondatasets}
\end{figure}

We collected data on the HM3D and MP3D datasets to build a text dataset for fine-tuning SBERT. HM-3D and MP-3D have annotated the regions where objects are located, as shown in the figure \ref{fig:regiondatasets}. We let the agent fully explore multiple scenes in the training set to obtain the true values of observable objects from the same frame camera perspective, forming scene descriptions. We used the SBERT base model to determine the scene to which the scene description belongs. If the object environment discrimination accuracy of a certain frame is greater than 80\%, it is a positive example; if it is less than 20\%, it is a negative example. Some sample pairs are shown in the table \ref{tab:sbertfinetune}.

\begin{table}[htbp]
    \centering
        \begin{tabular}{l|l}
        \hline
        \textbf{Anchor} & livingroom\\
        \textbf{Positive} & decorative\_plant tv tv stand door door mirror door frame door\\
        \textbf{Negative1} & bed decorative plant tv stand door door frame door\\
        \textbf{Negative2} & bed tv stand door frame door door mirror mirror\\
        \hline
        \textbf{Anchor} & kitchen\\
        \textbf{Positive} & sink kitchen\_cabinet bouquet window kitchen\_island ceiling\_lamp window\\
        \textbf{Negative1} & wall electronics bouquet window ceiling lamp window couch couch pillow\\
        \textbf{Negative2} & picture dining chair window decorative plant dining table mirror window stairs\\
        \hline
        \end{tabular}
    \caption{sbert finetune datasets}
    \label{tab:sbertfinetune}  
\end{table}

The training corpus comprises 15 environmental classes (\textit{bedroom, kitchen, bathroom}, etc.) derived from HM3D and MP3D datasets. Each sample contains an anchor scene description, positive example, and three negative examples. The preprocessing pipeline executes:

The semantic reasoning module employs \texttt{all-MiniLM-L6-v2} as the base architecture - a 6-layer transformer model with 384-dimensional embeddings, pretrained on semantic textual similarity tasks. We implement domain adaptation through triplet loss optimization, maintaining the original model's encoder structure while replacing the final projection layer to align with our environmental attribute classification objectives.

Training utilizes the \texttt{SentenceTransformers} framework with mixed-precision computation (\texttt{bf16=True}). Key hyperparameters include:
\begin{itemize}
    \item Learning rate: $2 \times 10^{-5}$ with linear warmup (10\% of training steps)
    \item Batch size: 16 samples per device using duplicate-free sampling
    \item Training epochs: 50 with early stopping based on validation loss
    \item Gradient clipping: Maximum norm of 0.2
    \item Optimization: AdamW with $\beta_1=0.9$, $\beta_2=0.999$
    \item Gradient clipping: Maximum norm $\|\nabla\|_2 \leq 0.2$
\end{itemize}

\begin{figure}[htbp]
    \centering
    \includegraphics[width=1\linewidth]{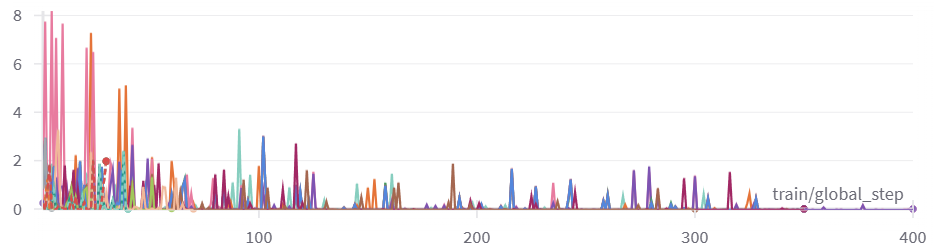}
    \caption{Training Loss}
    \label{fig:tloss}
\end{figure}

\begin{figure}[htbp]
    \centering
    \includegraphics[width=1\linewidth]{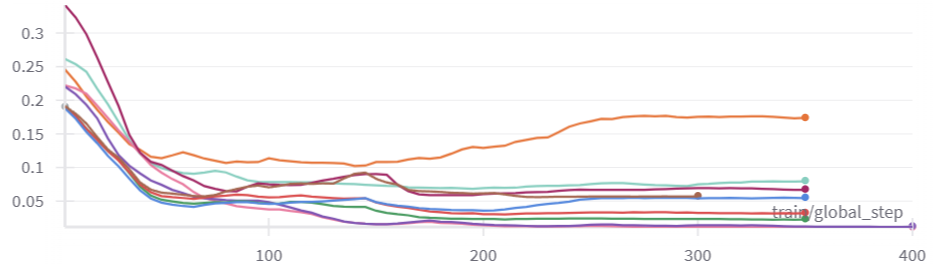}
    \caption{Evaluation Loss}
    \label{fig:eloss}
\end{figure}

Training loss and evaluation loss are shown in the figure \ref{fig:tloss} and figure \ref{fig:eloss}. Model checkpoints are saved every 5 evaluation steps ($\sim$250 iterations) with best weights selected by validation triplet accuracy. The final model demonstrates 51.1\% scene classification accuracy on HM3D validation split, representing a 12.3\% absolute improvement over the base pretrained model.

\subsection{MLLM Configurations}

\begin{figure}
    \centering
    \includegraphics[width=1\linewidth]{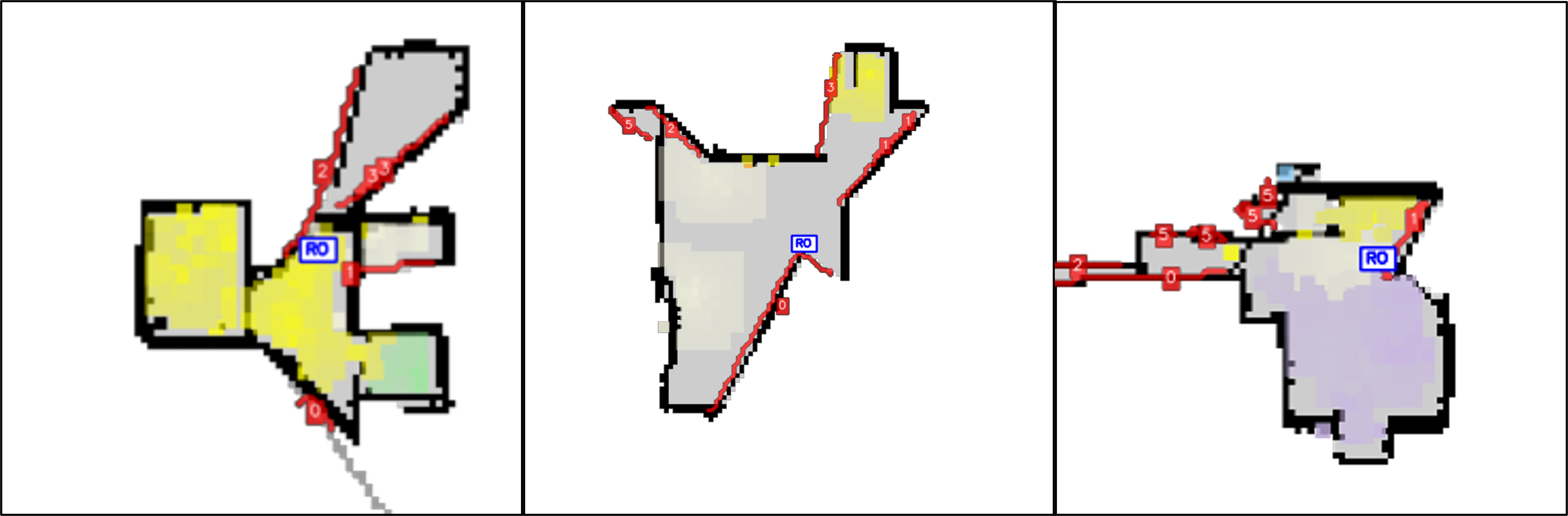}
    \caption{MLLM Figure Inputs}
    \label{fig:MLLMinput}
\end{figure}

\begin{figure}[htbp]
    \centering
    \includegraphics[width=1\linewidth]{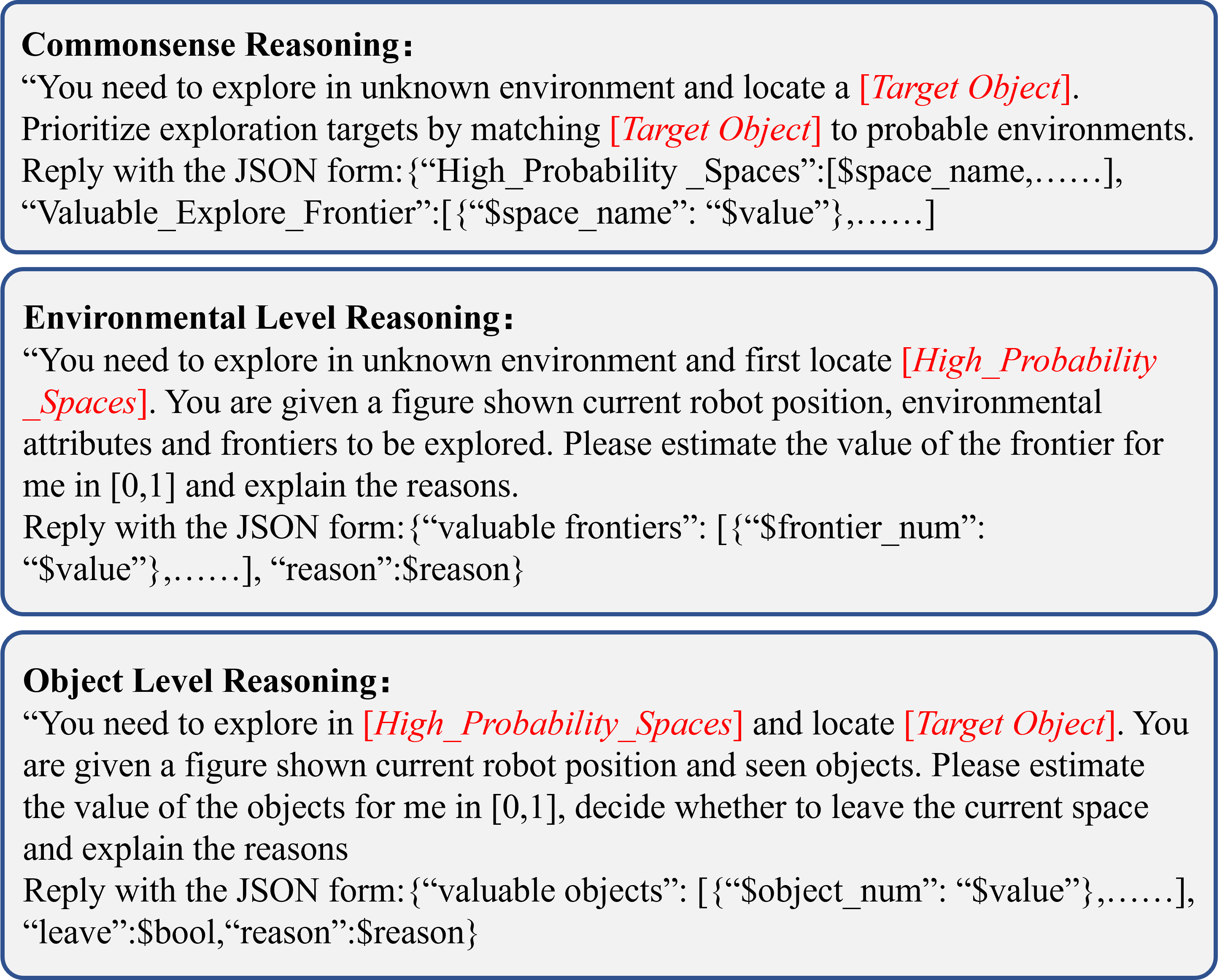}
    \caption{MLLM Prompt Template}
    \label{fig:prompt}
\end{figure}

We use Doubao-vision-pro-32k as the base model for multimodal hierarchical reasoning. The input for multimodal reasoning is shown in the figure \ref{fig:MLLMinput}. The prompt template is presented in the figure \ref{fig:prompt}.

\end{document}